\pdfoutput=1
\documentclass[11pt]{article}

\usepackage[preprint]{acl}

\usepackage{times}
\usepackage{latexsym}
\usepackage{graphicx}

\usepackage[T1]{fontenc}
\usepackage[utf8]{inputenc}

\usepackage{microtype}

\usepackage{inconsolata}

\usepackage{booktabs}
\usepackage{array}
\usepackage{longtable}
\usepackage{amsmath, amssymb}
\usepackage{listings}
\usepackage{tcolorbox}
\usepackage{subfigure}

\title{In-Memory Learning: A Declarative Learning Framework for \\ Large Language Models}

\newcommand*\samethanks[1][\value{footnote}]{\footnotemark[#1]}

\author{
Bo Wang\thanks{\ \ \ Equal contribution.}\thanks{Work done during internship at Shanghai Artificial Intelligence Laboratory}\quad\quad
Tianxiang Sun\samethanks\quad\quad
Hang Yan\quad\quad  \\
{\bf Siyin Wang}\quad\quad
{\bf Qingyuan Cheng}\quad\quad
{\bf Xipeng Qiu}\thanks{\ \ \ Corresponding author.}\\
School of Computer Science, Fudan University\\
Shanghai Artificial Intelligence Laboratory\\
\texttt{\{txsun19, siyinwang20, xpqiu\}@fudan.edu.cn}\quad
\texttt{\{bwang22, chengqy21\}@m.fudan.edu.cn}\quad \quad \\
\texttt{yanhang@pjlab.org.cn}\quad \quad
}

\begin{document}
\maketitle
\begin{abstract}
% Declarative learning is crucial for intelligent organisms to adapt to new environments, which can also apply to agents. In this paper, we explore the topic of whether agents can improve themselves without human-labeled data, leveraging approaches akin to declarative learning. We explore the key features of the benchmarks that evaluate the self-improvement process and introduce a novel learning framework, which we have termed In-memory Learning. Through systematic experimentation, we demonstrate the effectiveness of our framework and provide insights into this problem.
The exploration of whether agents can align with their environment without relying on human-labeled data presents an intriguing research topic. Drawing inspiration from the alignment process observed in intelligent organisms, where declarative memory plays a pivotal role in summarizing past experiences, we propose a novel learning framework. The agents adeptly distill insights from past experiences, refining and updating existing notes to enhance their performance in the environment. This entire process transpires within the memory components and is implemented through natural language, so we character this framework as In-memory Learning. We also delve into the key features of benchmarks designed to evaluate the self-improvement process. Through systematic experiments, we demonstrate the effectiveness of our framework and provide insights into this problem.
\end{abstract}

% 有关非陈述性学习的描述：an example of non-declarative learning. It demonstrates that individuals can learn to differentiate between various pitches in music through practice, yet they may find it challenging to articulate the process or method behind this skill.
\begin{figure*}[ht]
    \centering
    \includegraphics[scale=0.32]{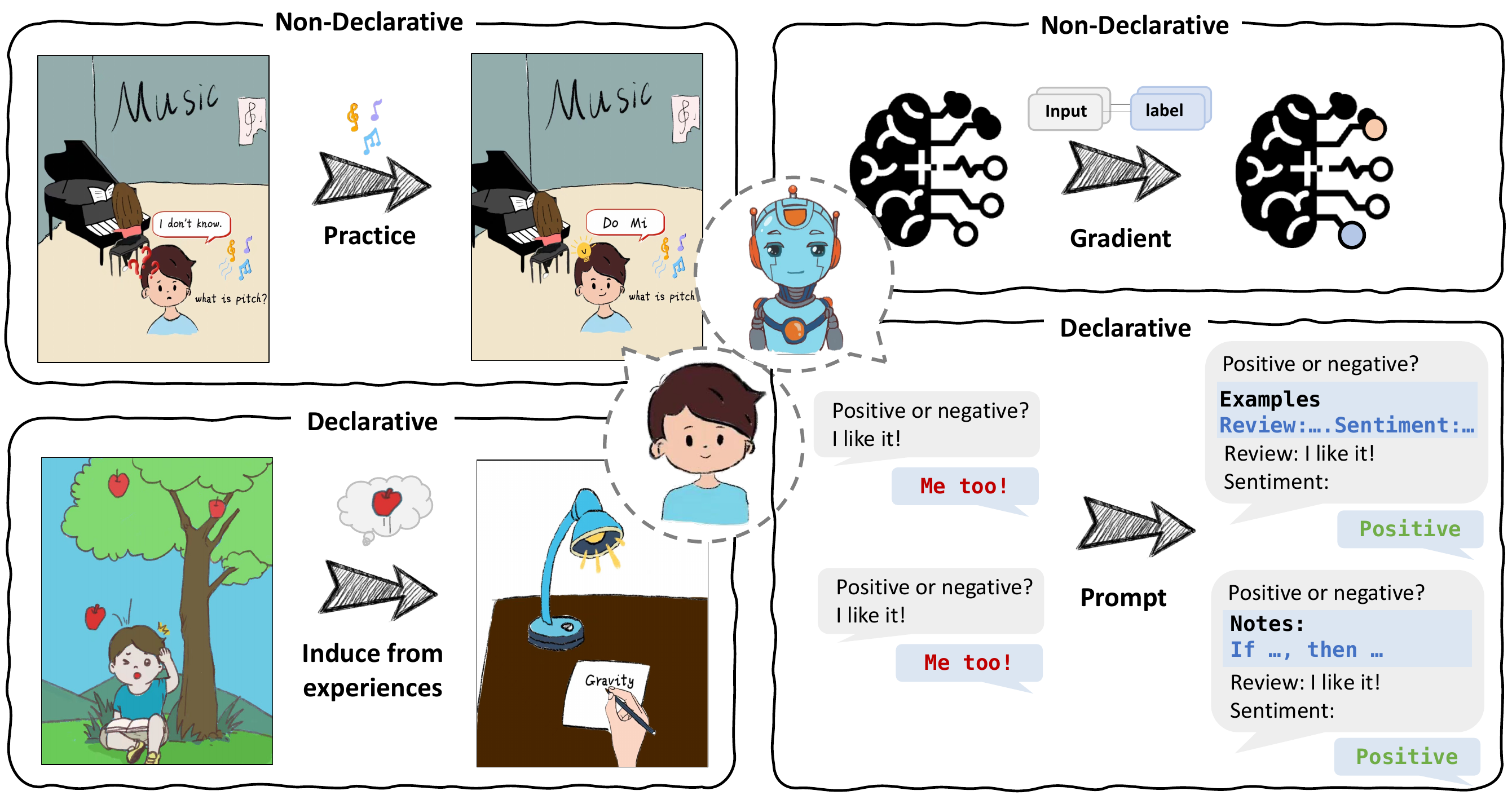}
    \caption{Learning Pattern. Non-declarative learning, as illustrated by the left figure, involves skills such as distinguishing relative pitches in music through practice. It's a challenge to express verbally. In contrast, declarative learning, exemplified by the right figure, refers to the acquisition of knowledge that can be explicitly stated, such as the introduction of the law of universal gravitation. For neural networks, models can develop the capability to answer questions through a gradient-based approach, as well as complete specific tasks using carefully designed prompts. This process closely resembles the learning process shown in the left parts.}
    \label{fig:learning pattern}
\end{figure*}

\section{Introduction}
The essential means by which intelligent organisms align themselves with changing environments is through learning and memory, which can be categorized into two distinct types in Neuroscience: \emph{declarative} and \emph{non-declarative} \citep{squire1996structure}. The memory acquired through non-declarative means is difficult to express in language, as depicted in Figure \ref{fig:learning pattern}. Conversely, declarative memory empowers individuals to convey past experiences with language, thus preparing them to navigate a wider array of scenarios with greater flexibility. When approaching new tasks or environments, humans summarize rules from initial experiences, subsequently refining and applying these rules to similar problems. This iterative refinement enhances understanding and effectiveness, gradually increasing familiarity with the task or environment.

When comes to Deep Neural Networks, if we liken learning through gradient back-propagation to a form of non-declarative learning, it can be observed that large language models \citep{brown2020language} benefit from an explicit formulation of their context window. Whether it involves generating the thought process using a Chain of Thought \citep{wei2023chainofthought} approach or providing input-output pairs as examples via In-context learning \citep{dong2023survey}, large language models get similar improvement to those gained through gradient-based methods, reducing the loss value and enhancing their performance in downstream tasks. As shown in Figure \ref{fig:learning pattern}, this method mirrors declarative learning, where understanding context enhances the network's performance. By leveraging this unique characteristic, agents built upon large language models can comprehend their environment, plan, and make decisions based on organizational context \citep{Shridhar_Yuan_Côté_Bisk_Trischler_Hausknecht_2020, xi2023rise}. This approach enables them to tackle a broad spectrum of problems effectively, which attracts the interest of many researchers.

Given that LLM-based agents exhibit capabilities similar to intelligent organisms, and recognizing that these abilities empower them to align with the natural world and enhance cognition, a natural question arises: Can agents develop similar self-improvement capabilities? Research on the autonomous agent \citep{qin2023toolllm, schick2023toolformer} usually incorporates the use of tools to formulate their context window autonomously, including strategies for teaching agents to utilize these tools or the design of processes that involve tools \citep{wang2023voyager}, such as retrievers. The enhancement in agent performance is significantly influenced by the performance of these tools, which can not improve themselves concurrently. The central question we are concerned about is whether agents can self-enhance in the absence of human-labeled data, which is the inherent capability of the model itself.

In this research, we propose a novel perspective on the learning process of agents, drawing inspiration from declarative learning methods employed by humans. We introduce a comprehensive learning framework, termed \emph{In-Memory Learning} (IML), which encompasses three pivotal components: \emph{induction}, \emph{revision}, and \emph{inference}. The learning process is completed in the memory component, which is what the name refers to. In analogy to the gradient calculation process in gradient-based learning, agents perform note induction from their current experience to identify an update direction, subsequently updating their previous notes. Through iterative updates, the rules summarized by the agents progressively align to the correct direction. Our experiments illustrate that, through applying this framework, the model can self-enhance without the requirement for human-annotated labels. The successful implementation of this method necessitates three distinct capabilities: 
\begin{itemize}
    \item \textbf{Induction}: the distillation of general principles from current experiences
    \item \textbf{Revision}: the refinement of pre-existing guidelines
    \item \textbf{Inference}: the application of these updated rules for logical reasoning. 
\end{itemize}
It's worth noting that we do not directly compare our framework with those that incorporate tools within agent systems, as our objective is to demonstrate the inherent potential for agents to self-improve. Instead, we further delve into an analysis of the model's capabilities and the impact of various IML parameters. 

Our main contribution is:
\begin{itemize}
    \item We discuss the essential properties that a benchmark requires to evaluate self-improvement abilities and have implemented a preliminary version of such a benchmark.
    \item We introduced a novel framework named In-memory Learning and carried out a comprehensive series of systematic experiments to investigate its effectiveness and capabilities.
\end{itemize}

\section{Related Work}
\subsection{LLM-Agent}
Discussions about agents have erupted, given the capacity of large language models to tackle a variety of language tasks, as previously mentioned. A particularly intriguing question arises regarding the self-improvement of these agents. In numerous studies, agents have demonstrated the ability to leverage tools to enhance their performance \citep{Yao_Zhao_Yu_Du_Shafran_Narasimhan_Cao_2022, schick2023toolformer, qin2023toolllm, shen2023hugginggpt, karpas2022mrkl, li2023apibank}. In the Reflexion \citep{shinn2023reflexion} framework, the model takes multiple trials on the same question, necessitating specific conditions to determine the appropriate moment to stop attempts.

Similar to the Voyager \citep{wang2023voyager}, we believe that the agent should operate within a stable environment over a long period. In practical scenarios, where labels are hard to obtain, the agent must develop an understanding of its surroundings and enhance its capabilities, diverging from the traditional notion of an autonomous agent. We later developed the concept of 'lifelong agent' in Voyager, to which our methods are specifically tailored. It's worth noting that the common practice for agents based on retrievers directly is acquiring related experiences and integrating them into the context \citep{wang2023voyager}, which essentially is in-context learning. Consequently, we have selected in-context learning as our foundational baseline. ExpeL \cite{zhao2023expel} also explores a similar process. The primary distinction from our work is we focus on iterative improvement and conduct systematic experiments about it, while ExpeL primarily emphasizes the benefits of cross-task experience.

\subsection{Agent Benchmark}
Existing benchmarks for agents assess model capabilities across multiple dimensions, such as the ability to function as an agent \citep{liu2023agentbench}, the planning skills necessary to address real-world issues \cite{Shridhar_Yuan_Côté_Bisk_Trischler_Hausknecht_2020, yao2022webshop, fan2022minedojo, ahn2022i} and their ability to complete tasks iteratively \citep{mohanty2023transforming}. The methods used to assess agents' performance vary widely, encompassing human evaluation through interviews \cite{park2023generative, lin2023agentsims} and subjective assessments \citep{choi2023llms}. However, there is a lack of benchmarks specifically designed to directly evaluate the self-improvement ability of agents \cite{xi2023rise}. We will discuss the characteristics of such a benchmark in the next section, which form the basis of our proposal for a new benchmark to measure agents' progression.

\begin{figure}[tbp]
    \centering
    \includegraphics[scale=0.47]{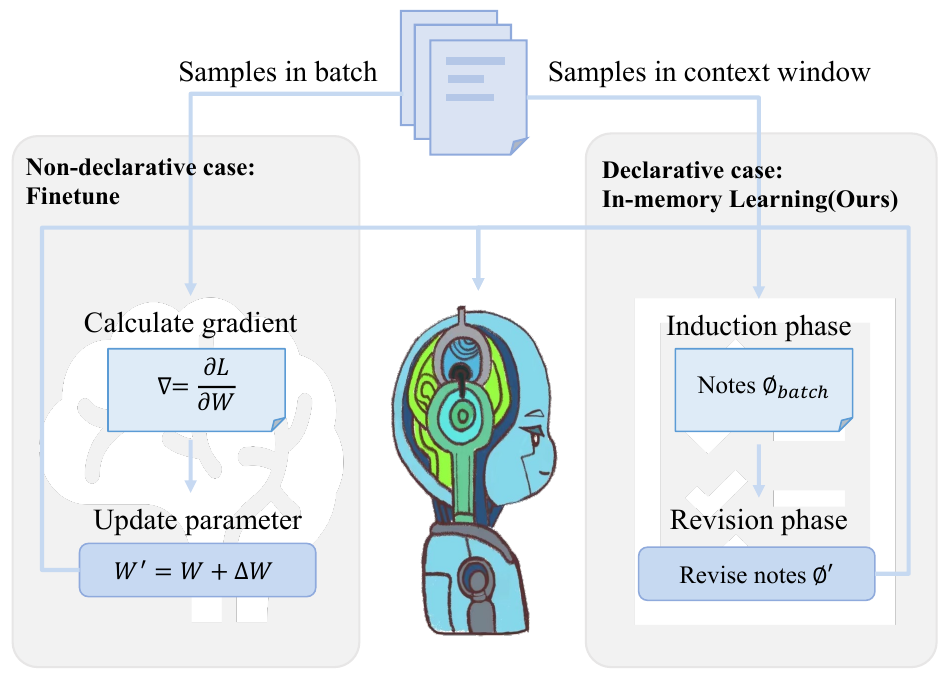}
    \caption{Backward Process. There is a similar structure between the gradient-based learning process and In-memory Learning(Ours)}
    \label{fig:backward process}
\end{figure}

\section{Meta Implementation}
The entire operation of an LLM-based agent can be formulated as a Partially Observed Markov Decision Process \citep{carta2023grounding} $(\mathcal{S}, \mathcal{V}, \mathcal{A}, \mathcal{T}, \mathcal{R}, \mathcal{G}, \mathcal{O}, \gamma)$ and we briefly introduce here. In this context, $\mathcal{S}$ is the state space while $\mathcal{V}$ represents the vocabulary of the language model. $A \subset \mathcal{V}^N$ is the action space and $\mathcal{G} \subset \mathcal{V}^N$ is the goal space. The transition function is represented by $\mathcal{T}: \mathcal{S} \times \mathcal{A} \mapsto \mathcal{S}$, the reward function by $\mathcal{R}: \mathcal{S} \times \mathcal{A} \times \mathcal{G} \mapsto \mathbb{R}$, and the observation function by $\mathcal{O}: \mathcal{S} \mapsto \mathcal{V}^N$.

Utilizing this definition, we can consequently define the problem of the Life-long Agent in section \ref{method: self-improved agent}, discuss the characteristics of the benchmark assessing the self-improve capabilities in section \ref{method:benmark}, and define the In-memory Learning Framework in section \ref{method: iml}.
\subsection{Self-improved Agent}
\begin{figure*}[htbp]
    \centering
    \includegraphics[scale=0.35]{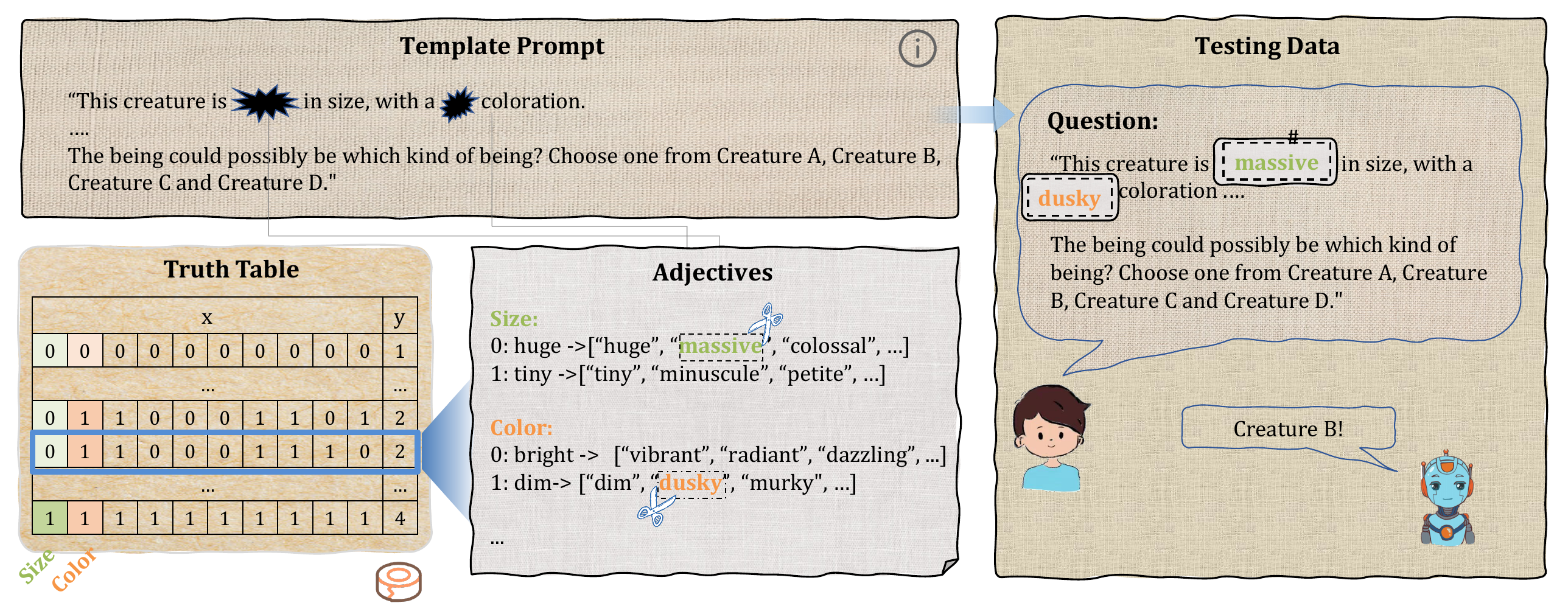}
    \caption{The construction process of our benchmark. We pre-define a correspondence from the \textbf{truth table} to the labels ($y$) and wrap it with natural language. Each column of the truth table represents a dimension of creatures ($x_i$), corresponding to two lists of \textbf{adjectives}. For instance, the first column stands for the size of the creature, associating the value 0 with huge and 1 with tiny. A combination of words is randomly selected from the sets of adjectives and then interconnected with predefined \textbf{prompts} to formulate the final \textbf{questions}.}
    \label{fig:benchmark}
\end{figure*}

\label{method: self-improved agent}
Agents in real-world scenarios are often tasked with consistently performing some specific types of tasks $\mathcal{G}_{spec} \subset \mathcal{G} \subset \mathcal{V}^N$ over an extended period. The question of the self-improved Agent centers on whether agents can enhance their performance without relying on human-labeled data since it's difficult to obtain such golden labels. Consequently, the reward function is categorized into two scenarios: one that utilizes fabricated labels such as AI feedback and the other in which only the correctness of outcomes can be known since it's often clear whether one solution has completed the task or not. In the implementation discussed below, we focus on the latter scenario.
\begin{equation}
\mathcal{R}\mapsto\left\{
\begin{array}{cl}
\mathbb{R}, & \text{fake labels exist}\\
\{0,1\}, & \text{else}
\end{array}
\right.
\end{equation}

where $\mathbb{R}$ on the right-hand side stands for the real set. The 'else' condition pertains to the correctness of the answer, 1 for correct and 0 for wrong.

\subsection{Benchmark}
\label{method:benmark}
The benchmark for assessing an agent's self-improvement ability should have certain essential characteristics. It should have a stable and clear testing goal to ensure that any progress by the model is noticeable. Additionally, the relationships within the data need to be learnable. Specifically, the least effective approach for self-improvement involves exhaustively searching through all possible solutions, which is meaningless here. Therefore, a relationship between the data is necessary. This also aligns with real-world scenarios, where common rules often exist across different experiences such as Newton's law of universal gravitation. Moreover, there must be enough data to make the problem statistically significant and solvable.

Since existing benchmarks are not designed to assess the ability for self-improvement, most of them do not fully align with the required features. For example, HotpotQA \citep{yang2018hotpotqa}, used in Reflexion, is primarily intended to evaluate multi-hop QA questions. However, upon analyzing errors made by agents that were tested by Exact Match(See Appendix\ref{sec:hotpotqa error}), we find that many of them are due to formatting issues, which are not expected and can't be generalized. As a result, we developed a straightforward classification dataset. We established a clear relationship between features and labels, making them learnable. The classification problem is suitably chosen because each correct feature-label match enhances the classifier's accuracy. The detailed information about the benchmark is introduced in Section \ref{exp: benchmark}.

\subsection{In-memory Learning}
\label{method: iml}
Within a Partially Observable Markov Decision Process (POMDP) trajectory $(s_0, o_0, a_0, s_1, r_1,  .., s_n, r_n)$, an agent selects an action based on $P(a|s, o, \theta)$, where $\theta$ represents all the variables, including prompts and parameters. Uniquely in our framework, we use the symbol $\phi$ to differentiate context notes from parameters of LLMs. The parameters of LLMs are frozen here and will therefore be omitted for simplicity. We will further explore the phases of the In-Memory Learning process in a formulaic manner below and introduce the details of implementation in section \ref{exp: impl detail}.

\subsubsection{Inference Phase}
In the inference phase, agents get the observation o about the current state $s$, and select an action $a \sim P(a|s, o, \phi)$. The reward r that the model receives aligns with the concept of the self-improved agent, which was mentioned before. The trajectory $\tau = (s_0, o_0, a_0, s_1, r_1)$ is recorded for later phase. This phase will continue until a specified threshold is reached.

\begin{figure*}[htbp]
    \centering
    \includegraphics[scale=0.5]{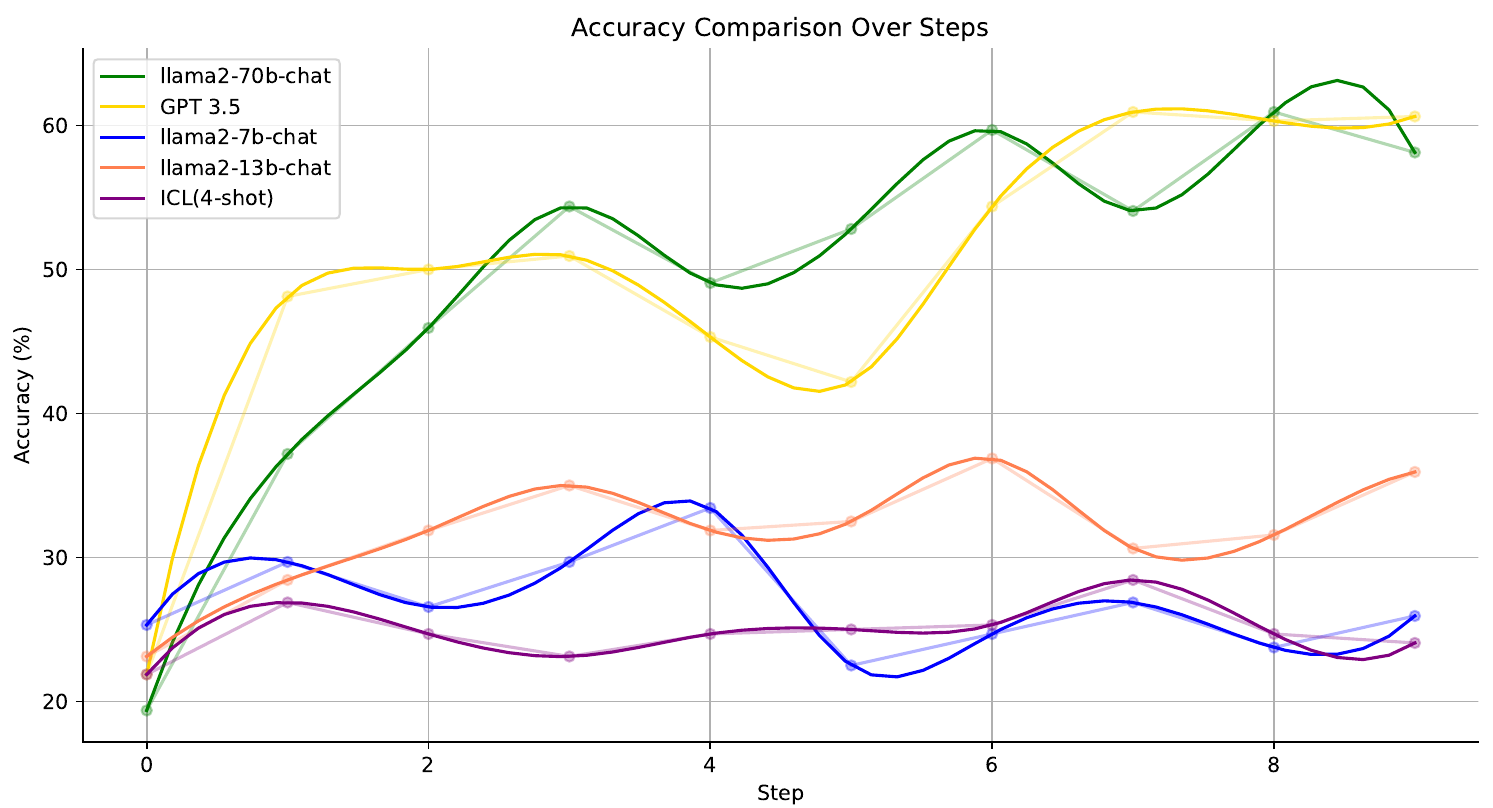}
    \caption{Accuracy curve over learning step. The solid lines represent the smoothed curves. Both \textbf{llama2-70b-chat} and \textbf{GPT-3.5-turbo} show an upward trend. \textbf{Llama2-13b-chat} also shows continuous improvement, but its performance is limited by its inference capabilities. \textbf{Llama2-7b-chat} initially improved but experienced a decline in later steps.}
    \label{fig:acc}
\end{figure*}

\subsubsection{Induction Phase}
After collecting a set of trajectories, the agent aims to derive general notes $\phi_{batch}$ from them. This process is completed using natural language descriptions, similar to calculating the gradient of batch data in gradient-based learning approaches like Figure \ref{fig:backward process}. The size of the batch for this inductive process is limited by the length of the context window, making the topic of long context windows particularly significant here.

\subsubsection{Revision Phase}
Like updating the parameter in gradient-based learning, the notes $\phi$ in the context before will be updated based on the insights $\phi_{batch}$ gained during the induction phase. The updated notes $\phi'$ will then be utilized in the subsequent inference phase. The correctness of updating direction is ensured by statistical properties, that common rules are consistent in different experiences.

\section{Experiments}
In this section, we will outline how we implemented the entire system first in section \ref{exp: impl detail} and carry out systematic experiments to evaluate its performance.
\subsection{Implementation Details}
\label{exp: impl detail}

\subsubsection{Benchmark}
\label{exp: benchmark}

To assess the self-improvement capabilities of agents, we developed a four-class classification problem. This problem involves a question describing one creature in 10 dimensions Like Figure \ref{fig:benchmark}, where every dimension is described by two opposing sets of adjectives. For instance, within the size dimension, one set of adjectives represents "huge" while the other represents "tiny". Each description uniquely matches a specific entry in a truth table that spans ten dimensions, thereby directly correlating to a single label.

In the real scenario, when hearing the name of a new species, some features can be inferred because the naming process often includes hints about its characteristics. So we use abstract labels, like "Creature A", to avoid bringing in this kind of prior information. For each entry of the truth table, four unique combinations of adjectives are randomly selected and 896 entries are held out for extension in the future. In the end, we get 3200 shuffled samples. The first two features are designed to be the distinguishing features while the others are distractors. The accuracy achieved on this task can significantly demonstrate the extent to which the agents have grasped these rules.
\begin{table*}[htbp]
    \centering
    \begin{tabular}{cccc}
    \toprule
    Model & Inference test(acc) & Induction test(acc) & Revise test ($\Delta$ acc) \\
    \midrule
    llama2-7b-chat & 37.11($\pm$ 9.46) & 43.31($\pm$ 5.02) & -3.81($\pm$ 12.36) \\
    llama2-13b-chat & 42.91($\pm$ 6.59) & 38.19($\pm$ 18.67) & 17.63($\pm$ 8.48) \\
    llama2-70b-chat & 58.67($\pm$ 9.51) & 48.44($\pm$ 6.3) & 1.063($\pm$ 5.09) \\
    GPT-3.5-turbo & 92.94($\pm$ 7.38) & 45.06($\pm$ 3.84) & 2.75($\pm$ 7.05) \\
    \bottomrule
    \end{tabular}
    \caption{Ability Test. The \textbf{inference test} applies five distinct formats of oracle notes to assess accuracy on the same test split. In \textbf{induction test}, agents summarize 80 groups of notes from the same 320 data samples. Using randomly sampled 5 groups to make inferences on the original 320 data samples and the same model. The \textbf{revision test} involves merging 5 pairs of notes into single notes. The accuracy differences are calculated between the minimum accuracy of pairs and their merged version.}
    \label{table:ability test}
    \extracolsep{\fill}
\end{table*}

\subsubsection{Inference Phase Implementation}
During the inference phase, the agent needs to identify which creature the description refers to. Initially, the notes $\phi$ are set to "no idea". A task-unrelated example is provided to guide the answering format of the agent and we use Exact Match to assess the accuracy of the agents' answers. By default, the agent processes 320 samples in a single step and saves the trajectories for use in the induction and revision phases. Following the implementation of Reflexion \cite{shinn2023reflexion}, we instruct the agent to respond with "Finish[Correct Answer]".

\subsubsection{Induction Phase Implementation}
After gathering trajectories in the previous phase, the agent identifies common features between them and summarizes their findings into batch notes $\phi_{batch}$. Due to the constraint of the context window, the induction phase is executed in minibatch while the results $\phi_{minibatch}$ are accumulated iteratively, summarizing into $\phi_{batch}$. We will delve into this process in the next section, demonstrating how such accumulation enhances stability, mirroring the effect of momentum observed in gradient-based learning. The notes are summarized for each creature individually and are later combined in the revision phase.

\subsubsection{Revision Phase Implementation}
Ultimately, the context notes for each creature are individually adjusted based on the batch notes and are then merged. We illustrate how the degree to which your instructions prompt the agent to make changes can impact the stability of the optimization process, similar to the momentum in gradient-based learning. Both the induction and revision phases occur within the agents' memory, leading us to name this approach as In-memory Learning.

\begin{figure}[ht]
    \centering
    \begin{subfigure}
        \centering
        \includegraphics[scale=0.29]{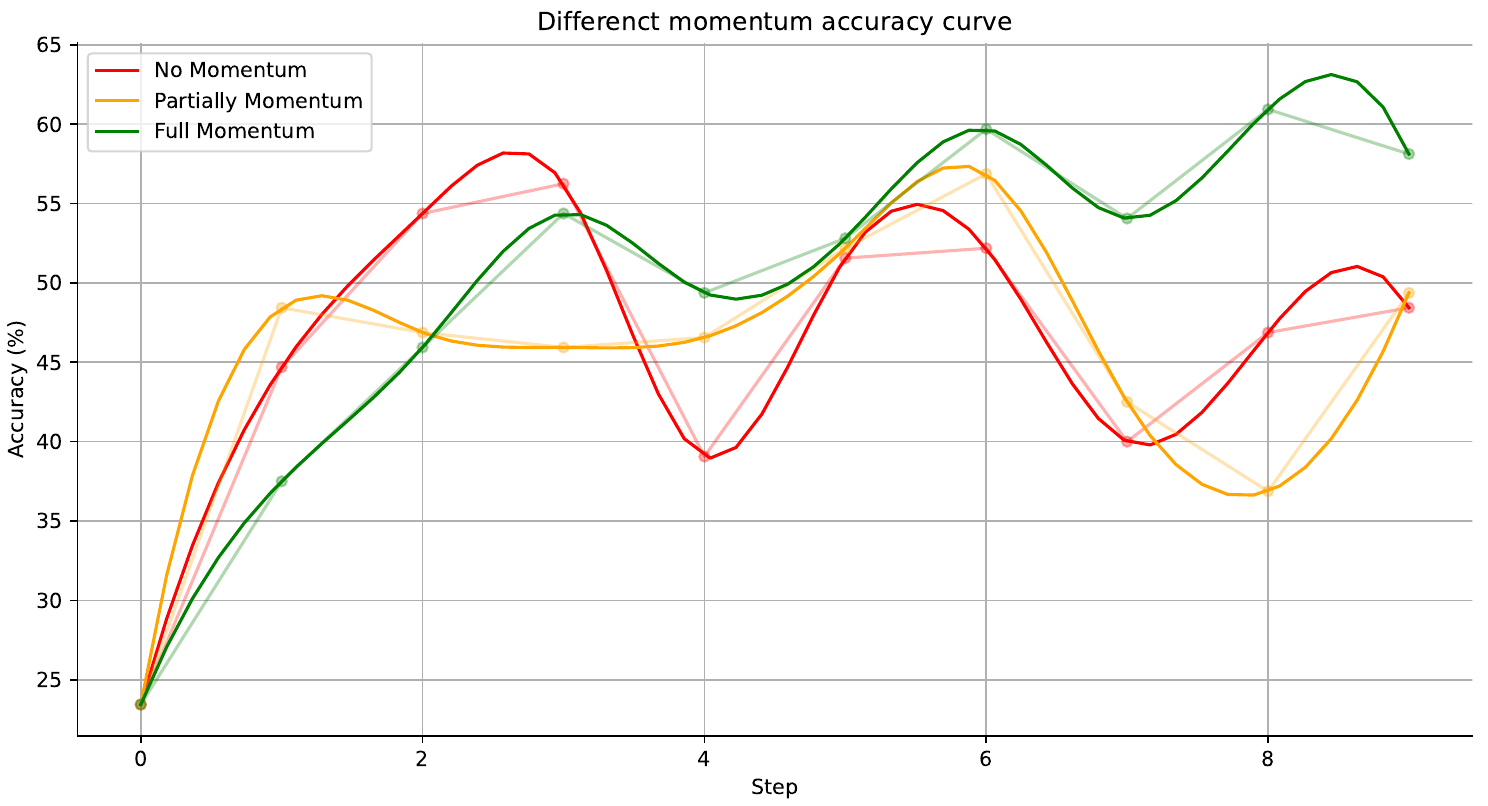}
        \caption{momentum test}
        \label{fig:momentum}
    \end{subfigure}
    \vspace{0.5cm}
    \begin{subfigure}
        \centering
        \includegraphics[scale=0.29]{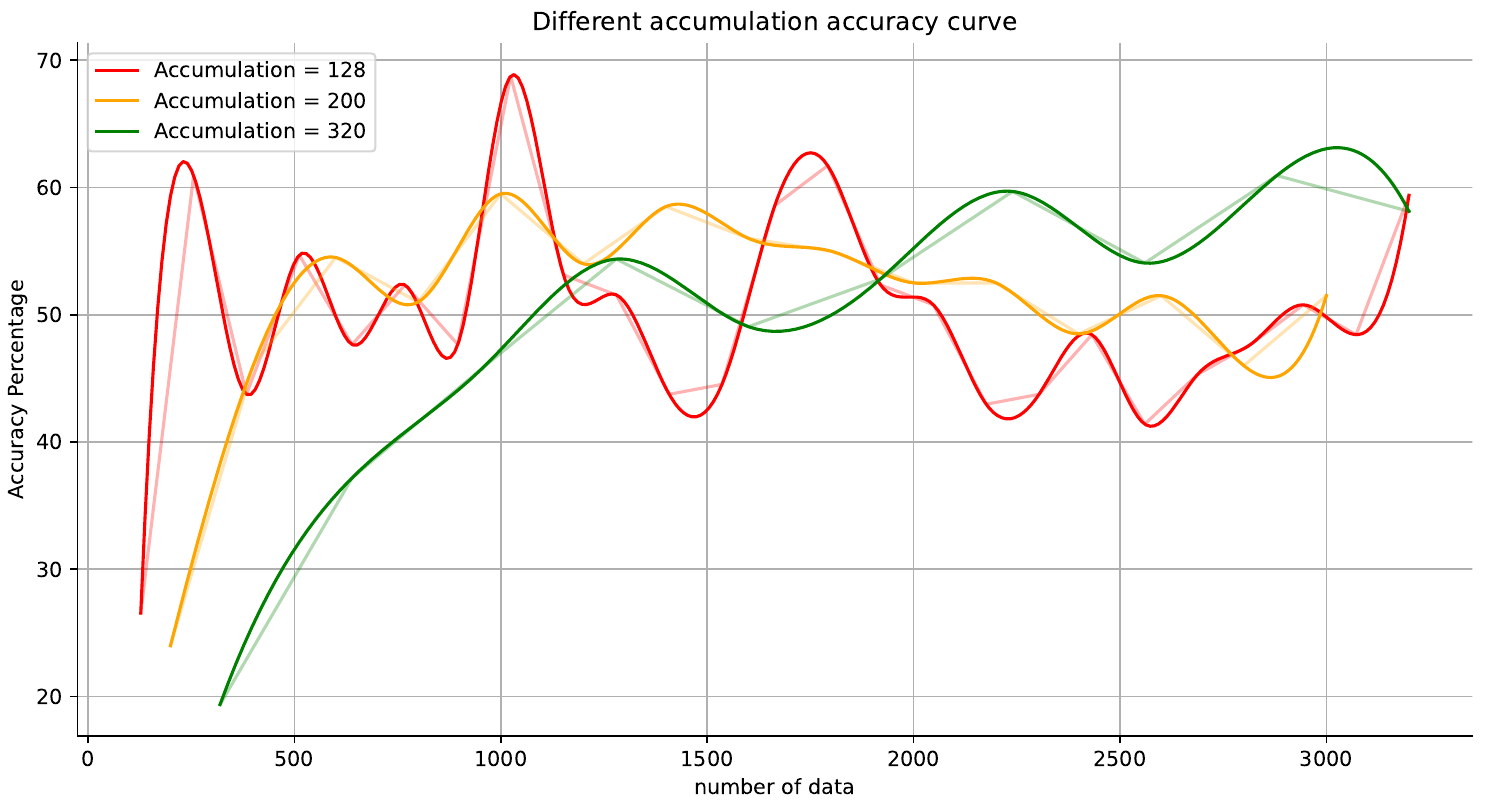}
        \caption{accumulation test}
        \label{fig:accumulation}
    \end{subfigure}
\end{figure}
\subsection{Compared with In-Context Leaning}
We choose In-context Learning as our baseline and the final result is presented in Figure \ref{fig:acc}. The result of in-context learning conducted in llama2-70b-chat is slightly better than random guessing. We use 4-shot as our benchmark consists of 4 labels, and the examples were manually chosen at random, ensuring the correctness of the answers. To validate the effectiveness of our approach, we conduct experiments using various models and analyze the outcomes.

\begin{figure*}[htbp]
    \centering
    \includegraphics[width=\textwidth, height=0.42\textwidth]{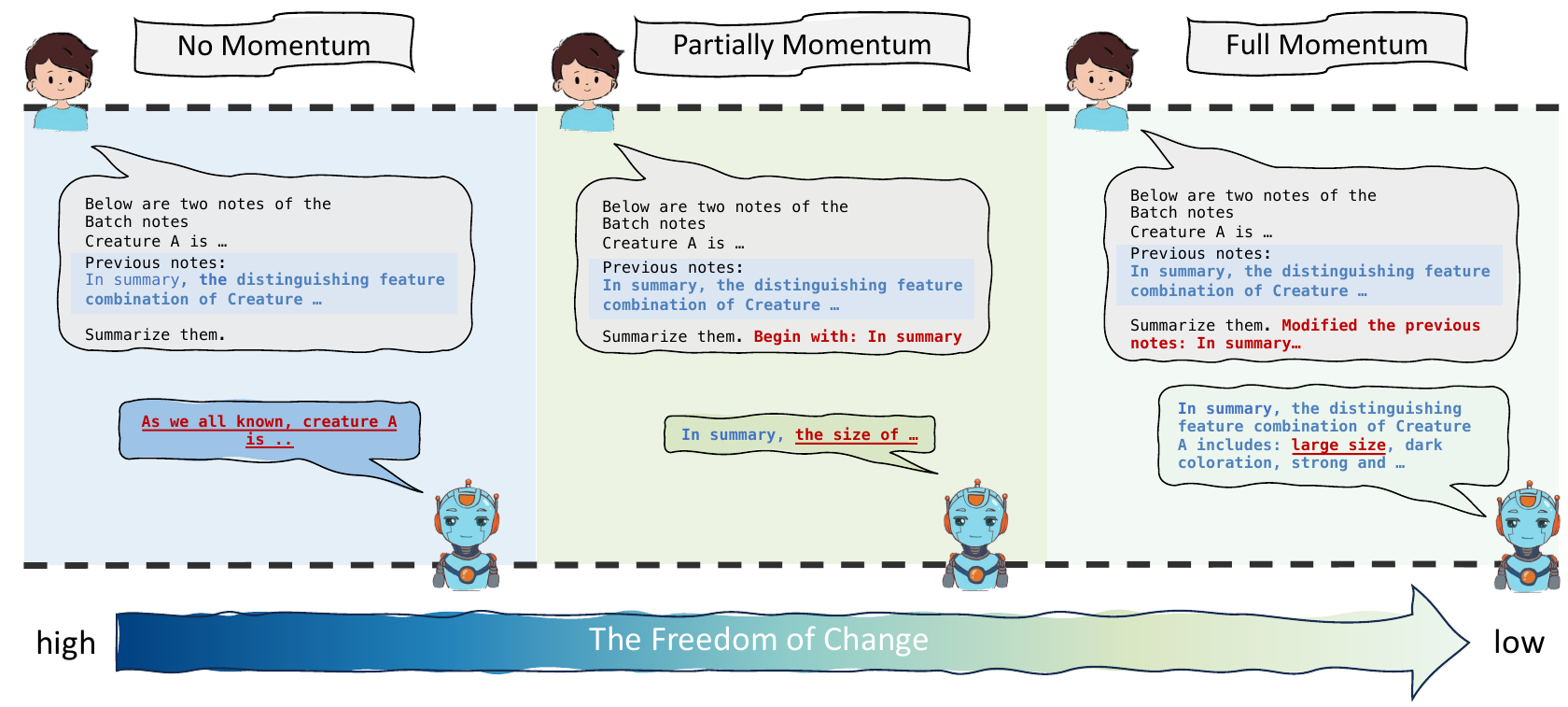}
    \caption{Momentum example. In the \textbf{No Momentum} setting, agents have the freedom to create new notes without any constraints. In the \textbf{Partially Momentum} setting, Agents are required to start with the initial words of the previous notes, which limits their freedom to make changes. The \textbf{Full Momentum} setting requires agents to make changes if necessary while appending the previous notes at the end of the prompts. The red underlined part in the reply represents the modified content compared to the previous notes.}
    \label{fig:momentum}
\end{figure*}

\subsection{Test on Various Models}
As depicted in Figure \ref{fig:acc}, the performance of GPT-3.5 and llama2-70b-chat shows a continuous improvement trend. However, llama2-13b-chat and llama2-7b-chat only improved a little and there is even a downward trend in the later steps for llama2-7b-chat. We analyze this outcome in three dimensions: the ability of inference, induction, and revision.

\subsubsection{Inference Ability}
We assess the inference ability of agents with Oracle notes, which indicate the upper bounds the agents can achieve in the inference phase. Given the sensitivity to the format of the prompt, we evaluate the accuracy of 5 different styles and compute the statistical result. The results shown in Table \ref{table:ability test} reveal that both the llama2-7b-chat and llama2-13b-chat models attain around 40 percent accuracy, explaining why the trend of improvement is not markedly evident, as the maximum accuracy with oracle notes is not high enough.

\subsubsection{Induction Ability}
The induction ability refers to the agent's capacity to summarize the common rules across different samples. In our study, four base models are tasked with performing induction on the same set of 320 samples, generating 80 groups of notes. We randomly select 5 of these 80 groups and use the llama2-70b-chat model to make inferences on the 320 samples. The results are presented in Table \ref{table:ability test}, indicating that llama2-70b-chat is the best one while llama2-13b-chat is the worst unexpectedly. The performance of GPT3.5-turbo falls short of that achieved by the llama2-70b-chat, providing insight into why GPT3.5 did not exhibit superior overall performance.

\subsubsection{Revision Ability}
During the revision phase, the agent is required to summarize two notes into one iteratively. To evaluate this capability, we devised a targeted experiment. Utilizing the notes collected by the llama2-70b-chat model, we randomly select 5 pairs of notes, and the agents need to merge each pair. We assess the agents' inference accuracy before and after the revision process. The difference in accuracy, that between the merged notes and the lower accuracy of the original pairs, serves as a measure of the agents' revision proficiency. The result is presented in Table \ref{table:ability test}. The llama2-7b-chat model exhibited a decrease in accuracy, which accounts for the model's declining performance in Figure \ref{fig:acc}. Conversely, the llama2-13b-chat model is the most superior one in this ability test.

\subsection{Effect of Parameters}
In our framework, certain key parameters influence the learning process. To explore these effects further, we conducted experiments focusing on the momentum and accumulation step, which are crucial for the stability of the learning process. We conduct the experiments on the llama2-70b-chat model.

\subsubsection{Effect of Momentum}
Although the natural language is discrete, our framework incorporates a momentum mechanism. As illustrated in Figure \ref{fig:momentum}, instructing the model to initiate responses using the initial words of previous notes acts as a form of momentum, constraining the generative freedom. Additionally, we incorporated basic statistical information regarding the quantity of samples processed by the agents. We conducted comparative analyses across different momentum settings, with the results shown in Figure \ref{fig:momentum}. In our experiments, the full momentum setting yields the most stable performance whereas the no momentum leads to the opposite. This suggests that integrating a momentum-like feature can significantly enhance the model's consistency.
\begin{figure}[htbp]
    \centering
    \includegraphics[width=0.48\textwidth]{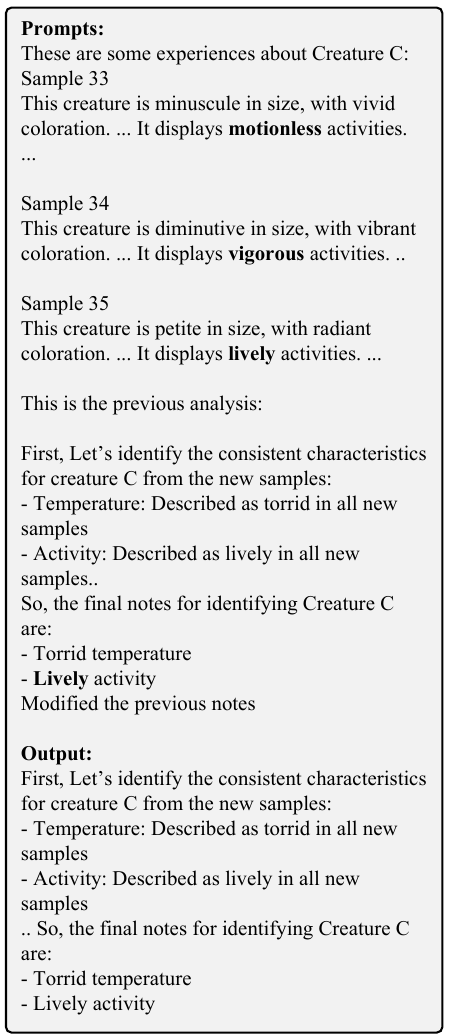}
    \caption{case study}
    \label{exp: local minimum}

\end{figure}

\subsubsection{Effect of Accumulation step}

Another critical parameter in our framework is the accumulation step count, which can exert influence on the learning process in two distinct ways. As described in the meta-implement section, the optimization process direction is determined by statistical properties, and the accumulation step assumes significance due to the fixed minibatch size imposed by the context window. Additionally, our assessments of accuracy during the subsequent influence phase are also influenced by the volume of data. In our experiment, we examined three accumulation step values: 128, 200, and 320, with the result presented in Figure \ref{fig:accumulation}. As observed, a smaller accumulation step leads to greater instability in the learning process.

\subsection{Trapped in Local Minimum}
An interesting observation about the learning process is the presence of optimization challenges analogous to the occurrence of saddle points in gradient-based learning. When tasked with modifying existing notes based on new experiences, the model may encounter difficulties in updating, even when the new experience contradicts the existing notes. This issue tends to occur more frequently in the intermediate and advanced stages of the iterative update step. Since we have observed this phenomenon across various models, including GPT-3.5-turbo, we believe that it's not solely attributed to the diversity of training data. Rather, it appears as if the copy mechanism of transformers is triggered with the end-of-sequence token remaining the most likely outcome after repeating the previous notes, even in the presence of changed experiences. We have not identified the minimum support set to delve deeper into this question and leave it for future exploration. Figure \ref{exp: local minimum} shows an simplified examples

% \begin{tcolorbox}[width=0.5\textwidth]
% \label{exp: local minimum}
% \textbf{Prompts}:\\
% These are some experiences about Creature C: \\
% Sample 33 \\
% This creature is minuscule in size, with vivid coloration. ... It displays \textbf{motionless} activities. ...\\

% Sample 34 \\
% This creature is diminutive in size, with vibrant coloration. ... It displays \textbf{vigorous} activities. ..\\

% Sample 35 \\
% This creature is petite in size, with radiant coloration. ... It displays \textbf{lively} activities. ...\\

% This is the previous analysis: \\
% First, Let's identify the consistent characteristics for creature C from the new samples: \\
% - Temperature: Described as torrid in all new samples\\
% - Activity: Described as lively in all new samples\\
% ..
% So, the final notes for identifying Creature C are:\\

% - Torrid temperature\\
% - Lively activity \\

% Modified the previous notes\\
% \textbf{Output}:\\
% First, Let's identify the consistent characteristics for creature C from the new samples: \\
% - Temperature: Described as torrid in all new samples\\
% - Activity: Described as lively in all new samples\\
% ..
% So, the final notes for identifying Creature C are:\\

% - Torrid temperature\\
% - Lively activity
% \end{tcolorbox}

\section{Conclusion}
In conclusion, we formally define the problem of self-improved agents. We discuss the key properties of a benchmark designed to evaluate agents' self-improvement capabilities and introduce a novel framework called In-memory Learning. Our systematic experiments demonstrate the effectiveness of this method and provide valuable insights into this domain.

\section*{Limitations}
Multimodality has the potential to incorporate richer information, which can enable agents more adaptable to complex situations. In our current work, we primarily focus on text and do not incorporate multi-modality situations. This aspect is left for future research.

Due to the constraint of budget, we didn't conduct experiments with GPT-4, leaving unanswered questions about its potential effectiveness as a learner and the extent of improvements it can achieve.

\bibliography{custom}
\newpage
\appendix

\section{HotpotQA Error Analysis}
\label{sec:hotpotqa error}
Below are some outputs of llama2-7b-chat on hotpotQA. We list some error data samples. A common mistake occurs when asking about the timing of an event, the model only responds with the year, whereas the standard answer includes the month or the date.
\lstset{
    language=Python, % 设置编程语言
    basicstyle=\ttfamily\small, % 设置基本样式
    breaklines=true, % 启用自动换行
    postbreak=\mbox{\textcolor{red}{$\hookrightarrow$}\space}, % 换行处标记
    showstringspaces=false, % 不特别显示字符串中的空格
}
\onecolumn
\begin{longtable}{p{\textwidth}} % 跨栏表格环境
% \centering % 居中表格
% 表格标题
% \label{tab:hall_of_fame} % 用于引用的标签
% \begin{tabularx}{\textwidth} % 使用tabularx自动调整列宽
\toprule
\textbf{Query:} Chicagoland Sports Hall of Fame was founded by the company located in what Washington town, near the state capital? \\
\midrule
\textbf{Supporting Article:} The Chicagoland Sports Hall of Fame, located in the Hawthorne Race Course, in Stickney/Cicero, near Chicago, Illinois, honors sports greats associated with the Chicago metropolitan area. It was founded in 1979 as a trailer owned by the Olympia Brewing Company parked at Soldier Field in Chicago. The Chicago Park District took over the exhibits in 1983. From 1988 the exhibits were displayed in Mike Ditka's restaurant until the restaurant closed in 1991. The Hall of Fame moved to the Maryville Academy in Des Plaines in 1996 and has operated under the guidance of Father John P. Smyth since that time. As of 2008, it was operating at Hawthorne. The Olympia Brewing Company was a brewery in the northwest United States, located in Tumwater, Washington, near Olympia. Founded in 1896 by Leopold Friederich Schmidt, it was bought by G. Heileman Brewing Company in 1983. Through a series of consolidations, it was acquired by Pabst Brewing Company in 1999; the Tumwater brewery was closed in 2003 but the Olympia brand continues, currently contract brewed by MillerCoors in southern California. \\
\midrule
\textbf{Answer:} Tumwater, Washington \\
\midrule
\textbf{Type:} Bridge \\
\midrule
\textbf{Level:} Hard \\
\midrule
\textbf{Model Prediction:} {Tumwater} \\
\midrule
\midrule
\textbf{Query:} Roger O. Egeberg was Assistant Secretary for Health and Scientific Affairs during the administration of a president that served during what years? \\
\midrule
\textbf{Supporting Article:} Roger Olaf Egeberg, M.D. (13 November 1902 – 13 September 1997 Washington, D.C.) was an American medical educator, administrator and advocate of public health. He was General Douglas MacArthur's personal physician during World War II in the Pacific theater. His other roles included Assistant Secretary for Health and Scientific Affairs in the Department of Health, Education, and Welfare (now the United States Department of Health and Human Services) during the Nixon administration and Dean of the University of Southern California's medical school.\textbackslash n\textbackslash n Richard Milhous Nixon (January 9, 1913 – April 22, 1994) was the 37th President of the United States from 1969 until 1974, when he resigned from office, the only U.S. president to do so. He had previously served as the 36th Vice President of the United States from 1953 to 1961, and prior to that as a U.S. Representative and also Senator from California. \\
\midrule
\textbf{Answer:} 1969 until 1974 \\
\midrule
\textbf{Type:} Bridge \\
\midrule
\textbf{Level:} Hard \\
\midrule
\textbf{Model Prediction:}1970s \\
\midrule
\midrule
\textbf{Query:} In regards to the high school that forced Manchester High School to change its name in 1922, when was its first graduating class? \\
\midrule
\textbf{Supporting Article:} Manchester High School Central is the oldest public high school in the state of New Hampshire. Located in the heart of Manchester, New Hampshire, approximately 2,100 students attend from communities such as Candia, Hooksett, and Manchester. The name was changed from Manchester High School in 1922 when Manchester West High School opened. Including Central, Manchester has a total of three public high schools.\textbackslash n\textbackslash n Manchester High School West, known as West High or West, is a public high school located in Manchester, New Hampshire. According to the \"Blue Book: Manchester High School West Hand Book \" (1940), the school was officially opened in September 1923 as a grade school and high school, hosting the seventh and eighth grades. In 1924, with an upsurge in enrollment to 136 pupils, the school became a dedicated high school. The first graduating class was in June 1925.\textbackslash n\textbackslash n Manchester High School West, known as West High or West, is a public high school located in Manchester, New Hampshire. According to the \"Blue Book: Manchester High School West Hand Book \" (1940), the school was officially opened in September 1923 as a grade school and high school, hosting the seventh and eighth grades. In 1924, with an upsurge in enrollment to 136 pupils, the school became a dedicated high school. The first graduating class was in June 1925. \\
\midrule
\textbf{Answer:} June 1925 \\
\midrule
\textbf{Type:} Bridge \\
\midrule
\textbf{Level:} Hard \\
\midrule
\textbf{Model Prediction:}1925 \\
\midrule
\midrule
\textbf{Query:} When was the British author who wrote the novel on which \"Here We Go Round the Mulberry Bush\" was based born? \\
\midrule
\textbf{Supporting Article:} Here We Go Round the Mulberry Bush is a 1967 British film made based on the novel of the same name by Hunter Davies. It was listed to compete at the 1968 Cannes Film Festival, but the festival was cancelled due to the events of May 1968 in France.\textbackslash n\textbackslash n Edward Hunter Davies, OBE (born 7 January 1936) is a British author, journalist and broadcaster. He is the author of a number of books, including the only authorised biography of the Beatles. \\
\midrule
\textbf{Answer:} 7 January 1936 \\
\midrule
\textbf{Type:} Bridge \\
\midrule
\textbf{Level:} Hard \\
\midrule
\textbf{Model Prediction:}1936 \\
\midrule
\midrule
\textbf{Query:} When was the track from which a sample was featured in \"Take Me to the Clouds Above\" released?\\
\midrule
\textbf{Supporting Article:} LMC are a British dance group consisting of producers, Lee Monteverde, Matt Cadman and Cris Nuttall. They have performed remixes for Scooter, Erasure, Dannii Minogue, Lasgo, Flip \& Fill, Robert Palmer and Shania Twain. LMC is best known for the track \"Take Me to the Clouds Above\" which featured a sample from \"How Will I Know\" by Whitney Houston, and \"With or Without You\" by U2 which topped the UK Singles Chart in early 2004, as well as going top 5 in Ireland and top 10 in Australia.\textbackslash n\textbackslash n \"With or Without You\" is a song by Irish rock band U2. It is the third track from their fifth studio album, \"The Joshua Tree\" (1987), and was released as the album's lead single on 16 March 1987. The song was the group's most successful single at the time, becoming their first number-one hit in both the United States and Canada by topping the \"Billboard\" Hot 100 for three weeks and the \"RPM\" national singles chart for one week, with a further three weeks at number two. \\
\midrule
\textbf{Answer:} 16 March 1987 \\
\midrule
\textbf{Type:} Bridge \\
\midrule
\textbf{Level:} Hard \\
\midrule
\textbf{Model Prediction:}1987 \\
\bottomrule
% \end{tabularx}
\end{longtable}

\end{document}